\documentclass[runningheads,a4paper]{llncs}

\usepackage{amssymb}
\usepackage{graphicx}

\usepackage{url}
\urldef{\mailsa}\path|igor.grabec@fs.uni-lj.si,|
\urldef{\mailsb}\path|eva.svegl@gmail.com,|
\urldef{\mailsc}\path|miha.sok@kclj.si|

\begin{document}

\mainmatter  % start of an individual contribution

\title{Development of a Sensory-Neural Network for Medical Diagnosing}

\titlerunning{Development of a Sensory-Neural Network for Medical Diagnosing}

\author{Igor Grabec$^1$, Eva {\v S}vegl$^2$ and Mihael Sok$^2$%
%\thanks{This research was was supported by the Slovenian Academy of Sciences and Art, Ljubljana and Amanova Ltd, Technology Park of Ljubljana, Slovenia}%
}
\authorrunning{I. Grabec, E. {\v Svegl}, M. Sok}

\institute{1 - Slovenian Academy of Sciences and Art, Ljubljana, Slovenia\\ 
2 - Faculty of Medicine, University of Ljubljana, Slovenia\\
\mailsa\\ 
\mailsb\\
\mailsc\\
%\url{http://www.springer.com/lncs}
}

\toctitle{Lecture Notes in Computer Science: 15ISNN}
\tocauthor{Authors' Instructions}
\maketitle

\begin{abstract}
Performance of a sensory-neural network developed for diagnosing of diseases is described. Information about patient's condition is provided by answers to the questionnaire. Questions correspond to sensors generating signals when patients acknowledge symptoms. These signals excite neurons in which characteristics of the diseases are represented by synaptic weights associated with indicators of symptoms. The disease corresponding to the most excited neuron is proposed as the result of diagnosing. Its reliability is estimated by the likelihood defined by the ratio of excitation of the most excited neuron and the complete neural network. 

\end{abstract}

\section{Introduction}
Medical diagnosing can be treated as a mapping of symptoms to characteristics of diseases \cite{Wag,Imp}. Our goal is to develop a sensory-neural network (SNN) by which this mapping could be performed automatically by a PC \cite{Sok,Sok2,Pro}. With this aim we first transform data about patient's condition into a proper form for the numerical processing. This transformation corresponds to sensing of symptoms by sensors at the input layer of the SNN shown in Fig.\,1. Similarly as data of patients, the characteristics of diseases are transformed and utilized to specify the synaptic weights of neurons in the next layer \cite{Gra}. At this specification we take into account the performance of a doctor at the diagnosis assessment. 
\begin{figure}
\centering
\includegraphics[height=0.33\textwidth, width=0.4\textwidth]{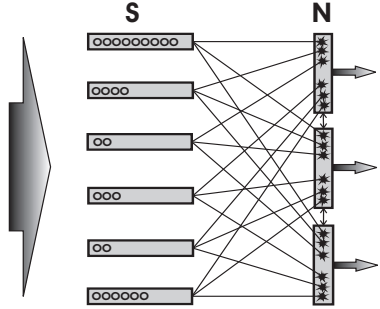}
\caption{Schematic drawing of the sensory-neural network. Bold arrows demonstrate input and output information flow, o - denote symptom indicators, ${\bf \ast}$ - show synapses,  $\leftrightarrow$ - represent connections between SNN elements.}
\label{F1}
\end{figure}
The doctor collects symptoms of the treated patient and compares them with the properties of diseases. At the comparison he considers certain symptoms as more significant than others and so assesses the agreement between given symptoms and those describing diseases \cite{Gra,Xia}. The disease with the highest correlation is then selected as a result of diagnosis. We describe such treatment by characterizing the significance of symptoms by synaptic weights connecting the sensors to neurons that correspond to various diseases. A signal from a particular sensor thus  contributes different amounts to excitation of various neurons. The excitation of a particular neuron therefore represents the correlation between the symptoms of a patient and the properties of the disease represented by the neuron. The disease corresponding to the most excited neuron in the NN layer determines the result of diagnosis. We describe the reliability of such assessment by expressing the output of the neuron relatively with respect to the mean value of all outputs. The corresponding estimator represents the likelihood of the diagnosis. At the specification of the SNN we first select a set of diseases and specify their symptoms as well as synaptic weights of neurons. In terms of them we next define the agreement and likelihood estimators.

\begin{table}[h]
\begin{center}
\caption{Symptoms used for characterization of diseases} \medskip
\begin{tabular}{|c|l|l|l|l|l|}  
\hline
\multicolumn{1}{|c|}{$s$} 
&\multicolumn{1}{l|}{Symptom name - $S(s)$} 
&\multicolumn{1}{c|}{$s$} 
&\multicolumn{1}{l|}{Symptom name - $S(s)$}
&\multicolumn{1}{c|}{$s$}
&\multicolumn{1}{l|}{Symptom name - $S(s)$} 
\\
\hline
1 & no symptoms& 17 & sleep disturb.- night walk& 33 & difficulty when speaking\\
2 & sudden onset of illness& 18 & distorted vision& 34 & difficulty moving one side\\
3 & hyperthermia& 19 & frequent urination& 35 & sudden infertility\\
4 & malaise, tiredness& 20 & tinnitus& 36 & difficult verbal expression\\
5 & age over 70& 21 & bleeding& 37 & vomiting\\
6 & headache& 22 & dark urine in last 2 weeks&  38 & snoring during sleep\\
7 & sudden headache& 23 & thirst, dry mouth& 39 & smelly urine\\
8 & cough& 24 & nervousness, depression& 40 & heart rate\\
9 & dizziness& 25 & throat irritation& 41 & arterial hypertension\\
10 & trouble breathing& 26 & pharyngeal dryness& 42 & pulse rhythmicity\\
11 & pain& 27 & hoarseness& 43 & O satur. - pulse oxim.\\
12 & w. loss in last 1/2 y& 28 & obstruction in the throat& 44 & breathing frequency\\
13 & loss of appetite& 29 & jaundice of skin \& sclera& 45 & body height\\
14 & irregular pulse& 30 & light feces& 46 & body weight\\
15 & sweating& 31 & frequent infections& & \\
16 & difficult concentrating& 32 & ability to lift both arms& &\\
\hline 
\end{tabular}
\end{center}
\label{Table 1}
\end{table} 
\begin{table}[h]
\begin{center}
\caption{Indicators of symptoms} \medskip
\begin{tabular}{|c|l|}  
\hline
\multicolumn{1}{|c|}{$s$}
&\multicolumn{1}{l|}{Indicators $I(s,:)$}   \\
\hline
3&\{fever, y, n\}\\  
6&\{y, n, in the morning, daily persistent\}\\  
8&\{y-in the morning, y-with sputum, y-daily persistent, y-without sputum,n\}\\  
10&\{at rest, on exertion, n\}\\  
11&\{chest,ear,abdomen,pharynx,at swallowing,at speaking,at urinating,shoulder,n\}\\ 
15&\{daily persistent, nocturnal,n\}\\
18&\{y, n, one eye only\}\\  
19&\{y, n, at night\}\\  
21&\{from the nose, in the sputum, in the urine, n\}\\  
31&\{urinary tract, tongue, corners of the mouth, lung, n\}\\  
40&\{$<$50, 51-70, 71-90, $>$90\}\\  
41&\{$<$100, 100-140, $>$140\}\\  
42&\{rhythmic, arrhythmic\}\\  
43&\{$\le$ 90, $>$90\}\\  
44&\{$<$10, 11-15, $>$15\}\\  
45&\{$<$150, 151-160, 161-170, 171-180, 181-190, $>$190\} cm\\  
46&\{$<$50, 51-60, 61-70, 71-80, 81-90, 91-100, $>$100\} kg\\  
\hline 
\end{tabular}
\\{\footnotesize
All other possible indicators are specified by: $I(s,:)=\{{\rm yes, no}\}=\{{ \rm y, n}\}$.\\
Components missing in Table 2 correspond to empty space $\emptyset$, for example: $I(s,:)=\{{ \rm fever, y, n}\}\equiv\{{\rm fever, y, n,\emptyset,\emptyset,\emptyset,\emptyset,\emptyset,\emptyset }\}$.}
\end{center}
\label{Table 2}
\end{table}
\section{Properties of data bases}
\subsection{Symptoms and diseases}
According to the suggestion of the research initiator \cite{Pro} we apply the set of $15$ most frequent diseases denoted by the name and index $D(d)$: 1) {\it Anemia}, 2) {\it Urinary tract infection}, 3) {\it Diabetes-2}, 4) {\it Atrial fibrillation}, 5) {\it Acute hemorrhagic stroke}, 6) {\it Obstructive sleep apnea}, 7) {\it Tuberculosis}, 8) {\it Chronic obstructive pulmonary disease}, 9) {\it Pneumonia}, 10) {\it Otitis media}, 11) {\it Leukocytosis}, 12) {\it Hepatitis-A}, 13) {\it Hypertension}, 14) {\it Throat inflammation}, 15) {\it No disease}. Their properties are characterized by $N_s=46$ symptoms $S(s)$ shown in Table 1 and used to describe the sensory layer of the SNN \cite{Sok,Sok2}. A symptom $S(s)$ is described by up to $N_i=9$ possible indicators $I(s,i)$ shown in Table 2. 

A particular disease is characterized by various symptom indicators. The presence or absence of a symptom indicator can be generally denoted by $0$ or $1$, but such indication leads to a very rough assessment of diagnosis. For a better assessment we describe the importance of the symptom indicator $I(s,i)$ of the disease $D(d)$ by a synaptic weight $W(d,s,i)$. A non important indicator is assigned the weight $W = 0$, while the other weights are positive or negative. A negative weight points to the absence of the disease. To avoid problems related with network training we apply a set of fixed weights specified by a clinical research. Their values mostly lie in the interval (1, 3). 
\begin{table}[h]
\begin{center}
\caption{Statistical weights describing importance of symptom indicators} \medskip
\begin{tabular}{|c|l|}  
\hline
\multicolumn{1}{|c|}{$d$} 
& \multicolumn{1}{l|}{The values denote: (symptom index $s$, indicator index $i$, weight $W(d,s,i)$)}\\
\hline
1 & (2,2,1) (3,3,1) (4,1,1) (6,1,1) (9,1,1) (10,1,1) (14,1,1) (16,1,1) (17,1,1) (20,1,1)\\ 
  & (31,1,1) (31,3,2) (40,4,1) (44,3,2)\\ 
2 & (2,1,1) (3,1,3) (3,2,1) (11,3,2) (11,8,4) (19,1,3) (22,1,3) (39,1,3)\\ 
3 & (2,2,2) (3,3,2) (4,1,1) (12,2,2) (18,1,1) (19,1,2) (31,1,2)\\ 
4 & (2,1,3) (3,2,3) (5,1,3) (10,2,2) (11,1,2) (14,1,3) (41,2,1) (41,3,3) (42,2,3)\\ 
5 & (2,1,2) (3,3,1) (7,1,2) (9,1,1) (18,3,2) (32,2,2) (33,1,2) (34,2,2) (35,1,2) (36,1,2)\\
  & (37,1,1)\\ 
6 & (2,2,1) (4,1,1) (6,3,1) (16,1,1) (17,1,3) (23,1,1) (24,1,2) (41,3,1) (46,6,2) (46,7,3)\\ 
7 & (2,2,1) (3,2,1) (4,1,1) (8,3,3) (11,1,1) (12,1,1) (13,1,1) (15,2,3) (21,2,3)\\ 
8 & (2,2,2) (3,3,1) (8,1,3) (8,2,3) (8,5,-3) (31,4,2) \\  
9 & (2,1,2) (3,1,3) (3,2,1) (4,1,2) (8,2,1) (8,3,1) (8,4,1) (10,1,1) (10,2,1) (11,1,1)\\ 
10 & (2,1,2) (3,2,1) (4,1,1) (6,4,2) (11,2,3) (29,1,1)\\ 
11 & (2,2,1) (3,2,1) (4,1,1) (9,1,1) (10,1,1) (10,2,1) (12,1,1) (13,1,1) (15,1,1) (18,1,1)\\ 
12 & (2,1,2) (3,2,1) (4,1,3) (11,3,2) (13,1,2) (29,1,3) 22,1,3) (30,1,3) (37,1,2)\\ 
13 & (2,2,1) (3,3,2) (5,1,1) (6,4,2) (9,1,1) (21,1,1) (41,3,6)\\ 
14 & (2,1,2) (3,1,1) (11,4,3) (25,1,2) (26,1,1) (27,1,1) (28,1,1)\\ 
15 & (1,1,1) (2,2,1) (3,3,1) (4,2,1) (6,2,1) (8,5,1) (11,9,1) (13,2,1) (16,2,1) (18,2,1)\\ 
   & (21,4,1) (22,2,1) (23,2,1) (24,2,1) (25,2,1) (26,2,1) (27,2,1) (28,2,1) (29,2,1)\\ 
   & (30,2,1) (31,5,1) (32,2,1) (33,2,1) (34,2,1) (35,2,1) (36,2,1) (37,2,1) (38,2,1)\\ 
   & (39,2,1)\\
\hline 
\end{tabular}
~\\{\footnotesize
Weights at other indices are $0$.}
\end{center}
\label{Table 3}
\end{table}
By the set $\{W(d,s,i); 1\le d \le 15; 1\le s \le 46; 1\le i\le 9\}$ shown in Table 3 we describe quantitatively the properties of diseases. For this purpose we treat the weights $W(d,s,i)$ as the transmission parameters of synaptic joints on the neuron with index $d$. To provide for equivalent treatment of all diseases at the assessment of a diagnosis, it is reasonable to normalize the weights so that the sum of their positive values equals $1$. By using the Heaviside function: $\{ H(x)=0$ for $x\le0$ ; $H(x)=1$ for $x>0\}$ we include just positive weights into the total positive weight: $W_t(d)= \sum_{s=1}^{N_s} \sum_{i=1}^{N_i}W(d,s,i) H(W(d,s,i))$, and define the normalized weight by the fraction: $W_n(d,s,i)=W(d,s,i)/ W_t(d)$.

\subsection{Algorithm of diagnosis} 
At a diagnosis, the patient first completes the questionnaire containing names of symptoms and their indicators. A confirmed item is represented by the value $1$ in the response matrix $\{ R(s,i); s=1\ldots N_s; i=1\ldots N_i\}$, while all other non-confirmed terms are $0$. This matrix represents the signals of values $0/1$ supplied from the sensory to neural layer of the network. We next assume that the signal from $I(s,i)$ excites the $neuron(d)$ over the synaptic weight $W_n(d,s,i)$ and adds to its output the amount $W_n(d,s,i)R(s,i)$. The complete output of the $neuron(d)$ is then given by the sum: 
\begin{equation}
A(d)=\sum_{s=1}^{N_s} \sum_{i=1}^{N_i} W_n(d,s,i) R(s,i)
\label{eq:agreement}
\end{equation}
The distribution $\{A(d); d=1,\ldots, N_d\}$ describes the excitation of the complete NN, while $A(d)$ is the estimator of the agreement between the patient's symptoms and the symptoms of the disease $D(d)$. If all symptoms of $D(d)$ coincide with the symptoms confirmed by the patient, the $neuron(d)$ is maximally excited and we get $A(d)=1$ (or $100\%$). As the result of the diagnosis assessment, we propose the disease $D(d_o)$ with the maximal agreement: $A(d_o)={\rm max}\, \{A(d);d=1\ldots N_d\}$. 

To support our decision we determine the mean value $<A>=\sum_{d=1}^{N_d}A(d)/N_d$ and the standard deviation $\sigma_A=\sqrt{{\rm var}(A)}$ of $A(d)$, and then determine the relative deviation: $\Delta_A(d) = (A(d)-<A>)/\sigma_A$. In the case when $\Delta_A(d_o)\gg 1$ we conclude that the corresponding $D(d_o)$ is significantly outstanding and can be considered as a reliable quantitative result of the diagnosis. Such reasoning is possible when the symptoms of a particular disease are well exhibited. However, this is not always the case, since several diseases share similar symptoms. This leads to similar stimulations and consequently, a question arises as how to infer when $\Delta_A(d_o)$ is not outstanding. To address this possibility we divide $A(d)$ by the sum: $ \sum_{d=1}^{N_d} A(d)$ and define the disease likelihood estimator $L(d)$ as:  
\begin{equation} 
L(d)=\frac{ A(d)} { \sum_{d'=1}^{N_d} A(d')}
\label{eq:likelihood}
\end{equation}
This estimator describes the relative stimulation of the $neuron(d)$ in the layer of neurons and represents the reliability of our decision. A high value of $L(d)$ corresponds to a firm argument for selecting the corresponding disease as a reliable result of the diagnosis. It is important that both $A(d)$ and $L(d)$ are quantitative in character and thus can be treated as the supplement of other quantitative data in clinical tests.    

\section{Performance of SNN}

\subsection{Characterization of the performance}

For a wide application we have developed a program that interacts with a user over a graphic interface \cite{Sok,Sok2}. At the start the graphic 
interface presents the user with three options: 1) Identifying a diagnosis based on 
completing the questionnaire, 2) Testing the program performance based on internally stored data sets of {\it formal symptoms} for all diseases, 3) Changing the weights of symptom indicators. All options can be repeated.

In the first option a new window with instructions for the user appears together with a window to enter the patient's name. After accepting the name, the program shows sequentially $46$ windows with the symptom names and their indicators. Confirmed data are translated into the response matrix $R$ that is led from the sensory to the neural layer of SNN, where the distributions of the agreement $A$ and likelihood $L$ are determined. The results are transferred to the 
user over various channels. The most informative is the displayed diagram of $A(d)$ and $L(d)$ distribution versus the disease index $d$. The lines at the levels of $< A >$, $<A>+\sigma_A$ and $<A>+2\sigma_A$ are used as references for a visual assessment of the diagnosis. Two  examples are shown in Fig.\,2. In addition to such diagram, the files with the patient's responses and corresponding numerical data are available for printing. 

In the second option the program displays the set of diseases. After one of them is selected, the program applies the corresponding formal set of symptoms and uses it instead of the patient's symptoms in the same procedure as in the first option. This step shows the optimal possibility of the selected disease diagnosis. 

The third option allows specialists to examine how a variation of synaptic weights influences the diagnosis process. By adjusting the weights and further testing the diagnosis using the second option, the performance of the complete program can be gradually improved. With this aim, the program allows modification of weights. This option provides for NN {\it training}, while the changing of the set of symptoms provides for {\it evolution} of the complete SNN. 

\begin{figure}
\centering
\includegraphics[width=0.49\textwidth]{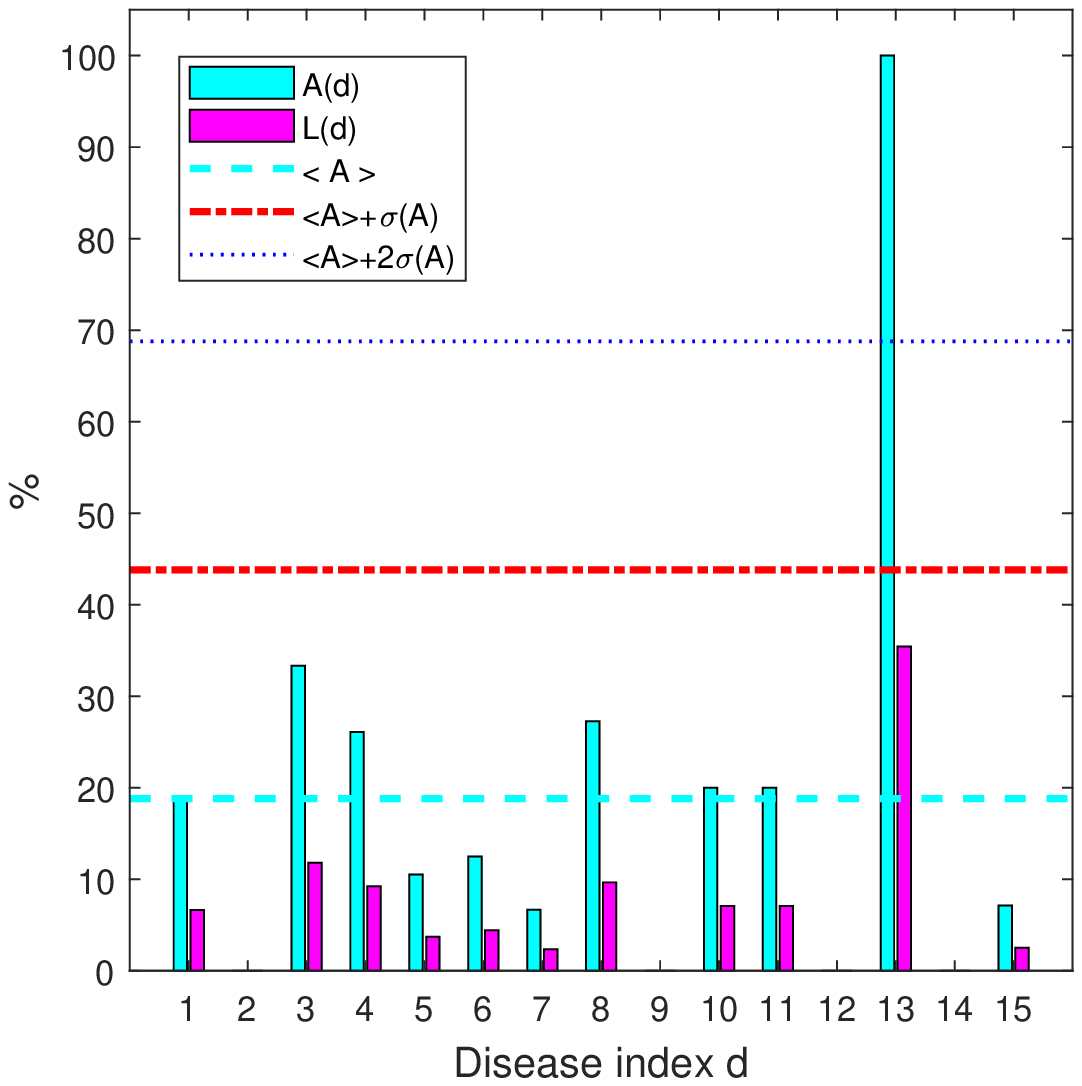}
\includegraphics[width=0.49\textwidth]{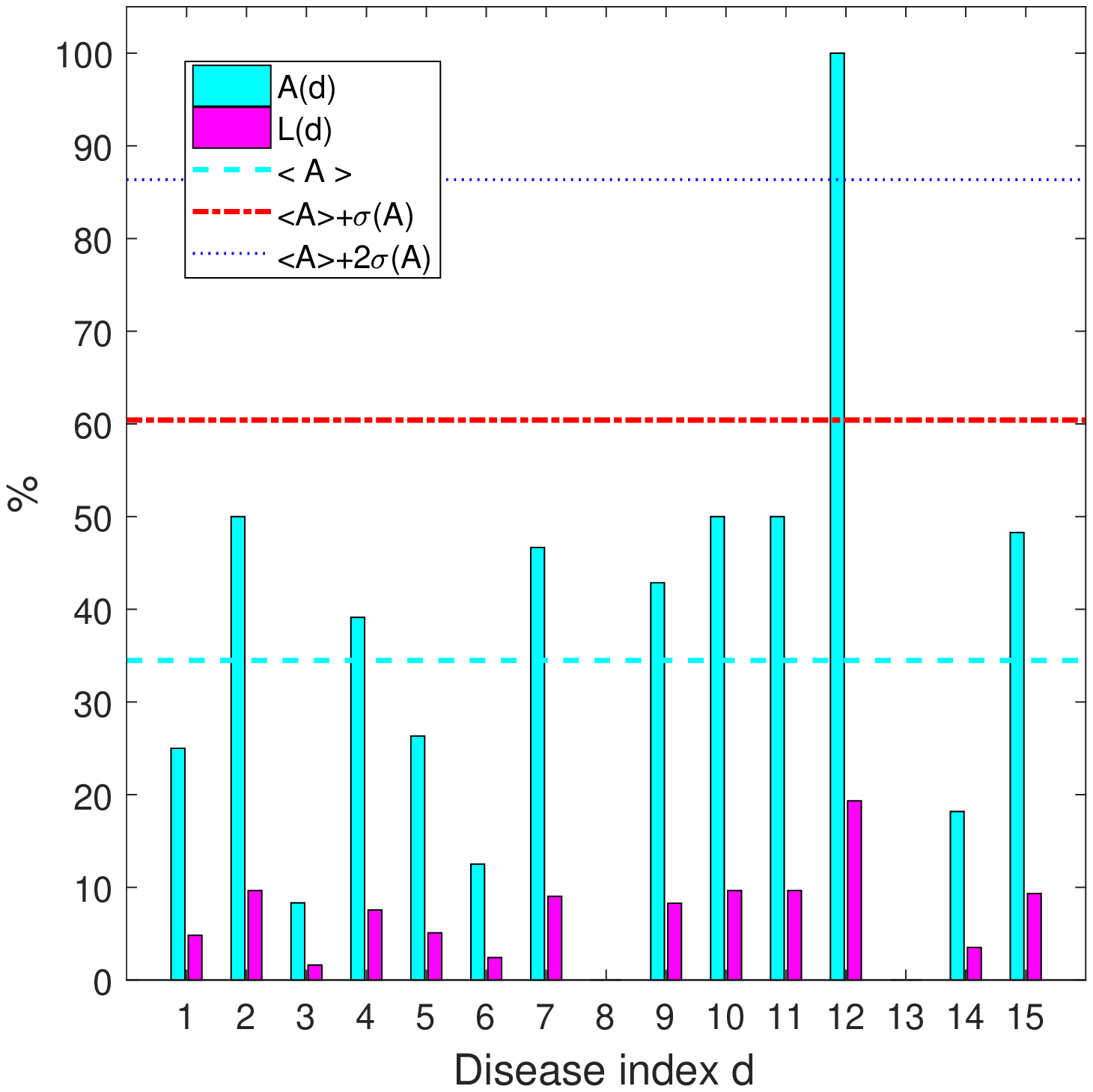}
\caption{Left : Distribution of $A$ and $L$ in testing the program performance based on formal symptoms of $D(13$) - {\it hypertension}. Right : Distribution of $A$ and $L$ based on the answers to the questionnaire by a patient suffering from $D(12)$ - {\it hepatitis A}. }
\label{F2}
\end{figure} 

\subsection{Testing of the program performance}
To demonstrate the performance of developed SNN we first present results of its testing performed with the formal symptoms of the typical disease $D(13)$ - {\it hypertension}. In this test, we get the diagram shown on the left of Fig.\,2. 

The value of the corresponding optimal agreement estimator is in this case $A_o(4)=100\%$, and it surpasses the mean value $<A>$ for $3.25\,\sigma_A$. This outstanding deviation from the mean value indicates a correct assessment of the diagnosis. But the  value of $A$ corresponding to several other diseases also surpasses the mean value $<A>$. This outcome indicates that the symptoms of these diseases are in a sense similar to those of {\it hypertension}. In spite of this property, the distribution of $A(d)$ suggests selecting {\it hypertension} as the result of the diagnosis. Although the agreement with its symptom indicators is $A_o=100\%$, the likelihood value of {\it hypertension} disease is only $L_o=35\%$; the other values, however, are still appreciably smaller: $L(d)\ll L_o ; d\neq d_o$. 

Similar performance of the program as in testing the diagnosis of {\it hypertension} is observed when using formal symptoms of other diseases. The corresponding optimal likelihood values $\{ L_o(d); d=1,\ldots 15\}$ are: 
\begin{equation}
L_o=(29,32,34,32,37,33,29,48,24,32,28,28,35,48,41)\%
\end{equation}
The mean value of the set of optimal values $\{L_o(d); d=1,\ldots ,15\}$ is similar as 
in the demonstrated case of {\it hypertension}, amounting to $<L_o>=34,6\%$. The corresponding standard deviation is $\sigma_L=7\%$ while the maximal and minimal values are $L_{o,max} = L_o(8) = L_o(14) = 48\%$ and $L_{o,min}=L_o(9)=24\%$, respectively. These data indicate that the likelihood value of $L_o\approx 35\%$ yields a rather firm quantitative argument for accepting the result of the automatic diagnosing. 

To demonstrate performance of SNN in the clinical practice we perform diagnosis of the disease $D(12)$ - {\it hepatitis A} by data obtained from a patient. The diagram is shown on the right of Fig.\,2. In this case the value of the relative agreement is also $A(12)=100\%$ and surpasses the mean value $<A>$ of the agreement distribution by $2.53\, \sigma_A$. All other values $\{A(d); d\not=12\}$ are essentially below the value $A(12)$; therefore, we accept as the rather reliable result of diagnosis the $D(12)$ - {\it hepatitis A}. This conclusion is supported by the value $L(12)=19\%$ that is smaller as $L_o(12)=28\%$ obtained at testing by formal indicators, but still approximately two times greater than all other values. Such result is obtained when all significant symptom indicators in the questionnaire are confirmed by a patient. Other characteristic examples are published elsewhere \cite{Sok,Sok2}.

\section{Conclusions}
Our testing has indicated that the selected sets of symptoms and synaptic weights provide a proper basis for the specification of a sensory-neural network applicable for an automatic diagnosing of selected diseases. The advantage of the developed method is the quantitative expression of the agreement between patient symptoms and properties of diseases. By using this estimator various subjective errors could be avoided at the assessment of a diagnosis. Moreover, its  reliability can be described by the disease likelihood estimator. Consequently, the developed SNN could be widely applied by medical doctors and patients outside the professional environment.  

At the development of our SNN we have utilized synaptic weights determined by doctors. However, in the interest of refining the diagnosing the corresponding data could also be automatically created and even improved during the application of the corresponding computer program. Various methods developed for training artificial neural networks could be applied for this purpose \cite{Wag,Gra}. Such an adaptation would in fact allow for the acquisition of new medical knowledge and also for its storage. 

We are aware that our procedure corresponds to a rather crude simplification of the professional performance of doctors. To improve it one should take more symptoms as well as diseases into account. However, making such an improvement requires more in depth descriptions of the corresponding sets. We expect that for this purpose applying a hierarchic structure could be advantageous.

\subsubsection*{Acknowledgments.} This research was was supported by the Slovenian Academy of Sciences and Art, Ljubljana and Amanova Ltd, Technology Park of Ljubljana, Slovenia.

\end{document}